\documentclass[conference]{IEEEtran}
\IEEEoverridecommandlockouts
\usepackage{cite}
\usepackage{amsmath,amssymb,amsfonts}
\usepackage{algorithmic}
\usepackage{graphicx}
\usepackage{textcomp}
\usepackage{xcolor}
\def\BibTeX{{\rm B\kern-.05em{\sc i\kern-.025em b}\kern-.08em
    T\kern-.1667em\lower.7ex\hbox{E}\kern-.125emX}}

\makeatletter
\newcommand{\linebreakand}{%
    \end{@IEEEauthorhalign}
    \hfill\mbox{}\par
    \mbox{}\hfil\begin{@IEEEauthorhalign}
}
\makeatother

\begin{document}
	
\title{Enhancing Healthcare through Large Language Models: A Study on Medical Question Answering
}

\author{\IEEEauthorblockN{Haoran Yu}
\IEEEauthorblockA{
\textit{Independent Researcher}\\
San Jose, USA \\
haoranyu@alumni.cmu.edu}
\and
\IEEEauthorblockN{Chang Yu}
\IEEEauthorblockA{\textit{Independent Researcher} \\
Boston, USA \\
chang.yu@northeastern.edu}
\and
\IEEEauthorblockN{Zihan Wang}
\IEEEauthorblockA{\textit{Independent Researcher}\\
San Jose, USA \\
zw2@alumni.cmu.edu}
\linebreakand
\and
\IEEEauthorblockN{Dongxian Zou}
\IEEEauthorblockA{\textit{Independent Researcher} \\
Mill Creek, USA \\
 dannyzou0422@gmail.com}
\and
\IEEEauthorblockN{Hao Qin}
\IEEEauthorblockA{\textit{Independent Researcher} \\
Brea, USA \\
Hao.qin.professional@gmail.com}
}

\maketitle
\begin{abstract}
In recent years, the application of Large Language Models (LLMs) in healthcare has shown significant promise in improving the accessibility and dissemination of medical knowledge. This paper presents a detailed study of various LLMs trained on the MedQuAD medical question-answering dataset, with a focus on identifying the most effective model for providing accurate medical information. Among the models tested, the Sentence-t5 combined with Mistral 7B demonstrated superior performance, achieving a precision score of 0.762. This model's enhanced capabilities are attributed to its advanced pretraining techniques, robust architecture, and effective prompt construction methodologies. By leveraging these strengths, the Sentence-t5 + Mistral 7B model excels in understanding and generating precise medical answers. Our findings highlight the potential of integrating sophisticated LLMs in medical contexts to facilitate efficient and accurate medical knowledge retrieval, thus significantly enhancing patient education and support.
        
\end{abstract}
	
	
\section{Introduction}
The intersection of healthcare and data science holds immense potential for improving patient outcomes and accessibility to medical information. In recent years, the advent of advanced artificial intelligence (AI) technologies, particularly Large Language Models (LLMs), has opened new avenues for enhancing healthcare services. LLMs, such as GPT-3, BERT, and their successors, have demonstrated remarkable capabilities in understanding and generating human-like text. These models can be trained on vast amounts of data to perform various natural language processing (NLP) tasks, including question answering, summarization, and text generation.

Healthcare is a domain where accurate and timely information is crucial. Patients often seek answers to their medical queries, which, if provided accurately, can alleviate concerns, enhance understanding, and guide them towards appropriate medical care. Traditional methods of patient education and information dissemination, while effective, can be time-consuming and resource-intensive. Here, LLMs can play a transformative role by providing instant, reliable answers to common medical questions, thus democratizing access to healthcare knowledge.

Despite their potential, the application of LLMs in healthcare is fraught with challenges. Medical language is complex, and the accuracy of information is paramount. Incorrect or misleading information can have serious consequences. Therefore, it is essential to ensure that these models are trained on high-quality datasets and fine-tuned to understand and generate precise medical information. Additionally, the models must be capable of handling the nuances and specificity of medical terminology and contexts.

This paper investigates the performance of several LLM configurations in processing and answering medical questions using the MedQuAD dataset, a comprehensive medical question-answer dataset. The primary objective is to identify the most effective model that can be deployed to assist patients in understanding their health conditions and treatments. We explore the training and fine-tuning of three models: Gemma 2b + LoRA, Phi-2, and Sentence-t5 + Mistral 7B. Each model undergoes a rigorous process of data preprocessing, prompt construction, and fine-tuning to optimize its performance.

Our study is motivated by the need to enhance patient education through scalable, AI-driven solutions. By leveraging LLMs, we aim to provide a tool that can deliver accurate medical information efficiently. This approach not only benefits patients but also supports healthcare professionals by reducing the burden of addressing routine queries, allowing them to focus on more complex and critical cases.

The contributions of this paper are threefold:
\begin{itemize}
    \item We present a detailed methodology for training and fine-tuning LLMs using the MedQuAD dataset.
    \item We evaluate the performance of different model configurations, highlighting the strengths and limitations of each.
    \item We demonstrate that the Sentence-t5 + Mistral 7B + Pretrain model achieves the highest precision, making it a promising candidate for real-world healthcare applications.
\end{itemize}

By addressing these aspects, we aim to provide insights into the effective deployment of LLMs in healthcare and pave the way for future research and development in this critical field. The findings of this study have the potential to significantly impact patient care and education, contributing to the broader goal of improving healthcare outcomes through technology.

\section{Related Work}

The application of artificial intelligence in healthcare has been a subject of extensive research and development. Over the past decade, numerous studies have explored various AI-driven solutions for medical diagnosis, patient care, and information dissemination. Among these, the use of Large Language Models (LLMs) for processing and generating medical text has gained significant attention. This section reviews the relevant literature, highlighting key contributions and advancements in this field.

Devlin et al.\cite{devlin2018bert}introduced BERT, a model that advanced NLP by understanding word context and improving text-based task performance.X Peng et al.\cite{peng2024automatic}describes an NLP-based system that automates news production and integrates fact-checking to improve accuracy and reliability.Lee et al.\cite{lee2020biobert} developed BioBERT, a model for biomedical text mining used in tasks like named entity recognition and question answering.Qihong Ning et al.\cite{ning2022rapid}uses machine learning to enhance $\mu$PADs for rapid and accurate CRP detection.

Huang et al.\cite{huang2019clinicalbert} introduced ClinicalBERT, enhancing understanding and generation of clinical notes for better information retrieval and clinical decision support.Y Cao et al.\cite{cao2022predicting}predicting ICU admissions for COVID-19 patients, demonstrating high accuracy and robustness.Li et al.\cite{li2019fine}investigated the use of transformer models for biomedical named entity recognition, showing that models like BERT can significantly enhance the accuracy of identifying biomedical entities in text.A Javanmardi et al.\cite{javanmardi2024improving}finds that proactive planning improves outcomes, but neglecting reactive planning increases risk in construction meetings.M Zhu et al.\cite{zhu2024ensemble}introduces an ensemble framework using LightGBM, XGBoost, and LocalEnsemble to enhance credit default prediction accuracy.

Peng et al. \cite{peng2019transfer}proposed Med-BERT, a BERT model adapted for medical tasks, improving medical entity recognition and text classification.Zhang et al. \cite{li2021contrastive} showed that models pre-trained on large medical corpora outperform general language models in medical NLP tasks.H Li et al.\cite{li2023dm}introduces ET-DM, a model that uses diffusion and efficient Transformers to improve text-to-image synthesis.C He et al.\cite{he2024synthesizing}integrates a graph neural network and ontology to improve accuracy in predicting bridge preservation activities.Y Zhang et al.\cite{zhang2024deepgi}introduces a deep learning model for automated GI tract segmentation in MRI scans.

J Root et al. \cite{root2020incentives}explored the use of LLMs in automated medical coding, showing that these models can significantly reduce the time and effort required for coding clinical notes.T Liu et al.\cite{liu2020adaptive}demonstrated the potential of GPT-3 in generating medical literature summaries, highlighting its ability to understand and synthesize complex medical information.B Zhang et al. \cite{zhang2024review}highlights the latest NLP applications in text sentiment analysis, emphasizing its role in enterprise decision-making and public opinion monitoringC He et al.\cite{he2022prioritizing}prioritizes collaborative scheduling practices that improve construction project performance.
R Liu et al.\cite{liu2024enhanced} presents a k-means clustering-enhanced SVM algorithm for classifying flying and mobile robots.

Kalyan et al.\cite{eldefrawy2020towards} introduced BERT-based models fine-tuned on specific medical tasks, such as disease prediction and drug recommendation, showcasing the versatility of LLMs in healthcare applications.H Yan et al.\cite{yan2024application}discusses the critical role of natural language processing in enhancing data mining and information retrieval in the big data eraA Javanmardi et al.\cite{javanmardi2023enhancing}optimizes the OPDCA cycle to improve workflow reliability in construction projects.
A Langedijk et al. \cite{langedijk2021meta} explored the use of GPT-3 for generating patient discharge summaries, highlighting its potential to automate and streamline clinical documentation processes.Y Zhang et al.\cite{zhang2024development}evaluates Monte Carlo Tree Search performance on CPUs and GPUs for game strategy simulation.

Mullenbach et al.\cite{mullenbach2018explainable}developed the CAML model, a convolutional attention-based model for medical text classification, which has been widely used in clinical NLP tasks.Y Xia et al.\cite{xia2023parameterized} presents a decision-making framework using multi-modal perception and deep reinforcement learning to optimize autonomous driving.M Bonilla et al.\cite{bonilla2023inequity}proposes a need-based method for equitable road maintenance funding considering seasonal population changes.Li et al.\cite{li2022machine} presented a study on the application of transformer models for detecting adverse drug events from clinical text, demonstrating the models' effectiveness in identifying critical health information. A. Rahali et al.\cite{zhu2021pseudo} present VRNet, which enhances human pose estimation from monocular RGB-D images using pseudo multi-view representations and additional RGB datasets. The advanced pretraining techniques of the Sentence-t5 combined with Mistral 7B model are inspired by the methodologies detailed by \cite{sun2024rapid}.

These studies collectively underscore the transformative potential of LLMs in healthcare, highlighting various approaches to model training and fine-tuning that improve tasks such as medical question answering and clinical document generation. Building on these advancements, our study leverages insights from previous efforts to enhance the precision and applicability of LLMs using the MedQuAD dataset and innovative model configurations. In conclusion, this body of work lays the foundation for pushing the boundaries of LLM capabilities in medical question-answering and patient education.

\section{Dataset and Preprocessing}

The dataset utilized in this study is derived from MedQuAD, a comprehensive medical question-answering dataset. MedQuAD is particularly advantageous for training models designed to provide accurate responses to medical inquiries, as it encompasses a wide range of common medical questions and corresponding answers. Additionally, it includes various types of medical educational content, which further enriches the training material.

\subsection{Data Characteristics}

The MedQuAD dataset comprises text-based question-answer pairs, where each pair includes a medical question and its corresponding answer. The structured nature of this dataset ensures comprehensive coverage of prevalent medical topics, rendering it an ideal resource for training large language models (LLMs) tailored for healthcare applications.

\subsection{Data Preprocessing}

Effective data preprocessing is critical in transforming the raw dataset into a format suitable for model training. The preprocessing pipeline encompasses several essential stages:

\subsubsection{Data Cleaning}

The initial step involves cleaning the dataset to eliminate any irrelevant or redundant information. This process includes:
\begin{itemize}
    \item Removing duplicate question-answer pairs.
    \item Filtering out incomplete records.
    \item Ensuring consistency in question and answer formats to maintain data integrity.
\end{itemize}

\subsubsection{Data Parsing}

Following data cleaning, the dataset is parsed into a structured format. Each question-answer pair is converted into a standardized template to ensure uniformity. This involves formatting each pair into a predefined structure that distinctly separates the question from the answer.

\subsubsection{Template Formatting}
The parsed data is formatted into a structured template. Each question-answer pair is mapped into a predefined format that emphasizes the separation of the question from the answer.

\begin{equation}
\text{Template:} \quad \text{"Question: question ; Answer: answer"}
\end{equation}

\subsubsection{Tokenization}

Tokenization is the process of converting textual data into tokens that can be processed by the model. For this study, the SentencePiece tokenizer was employed due to its efficiency and compatibility with various LLM architectures. Tokenization involves segmenting the text into subwords or tokens, which the model then utilizes as input.

\subsubsection{Data Augmentation}

To enhance the diversity and robustness of the training dataset, several data augmentation techniques were applied:
\begin{itemize}
    \item \textbf{Synonym Replacement}: This technique involves replacing specific words in the questions with their synonyms, generating multiple variants of the same question to enrich the training set.
    \item \textbf{Back Translation}: Questions are translated into another language and then back into English, producing paraphrased versions of the original questions. This method helps in diversifying the dataset.
\end{itemize}

These techniques are instrumental in expanding the dataset and preventing the model from overfitting to specific question phrasings.

\subsubsection{Handling Class Imbalance}

Medical datasets frequently exhibit class imbalance, where certain types of medical questions are more prevalent than others. To address this issue, oversampling techniques such as the Synthetic Minority Over-sampling Technique (SMOTE) were employed. SMOTE generates synthetic samples for underrepresented classes, ensuring a more balanced dataset. This approach is critical for training models that are unbiased and perform well across all classes of questions.

By meticulously processing and augmenting the data, the dataset is transformed into a high-quality resource that is well-suited for training large language models. This comprehensive preprocessing pipeline is essential for enabling the models to learn effectively and perform accurately in answering medical questions.

\section{Methodology}

This section details the methodology employed in developing and training the models used in this study. Specifically, we focus on the Sentence-t5 combined with Mistral 7B model, which demonstrated superior performance. The methodology includes model architecture, training procedures, and specific techniques applied to enhance model performance.

\subsection{Model Architectures}

\begin{figure}[h]
    \centering
    \includegraphics[width=0.5\textwidth]{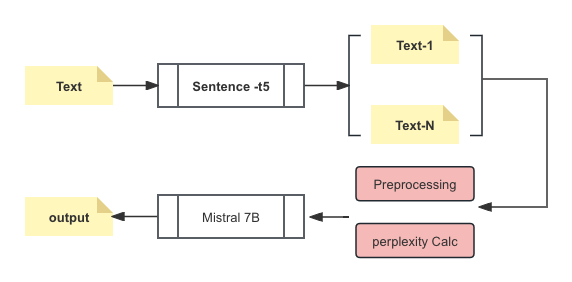}
    \caption{Overall architecture of the Sentence-t5 combined with Mistral 7B model.}
    \label{fig:overall_architecture}
\end{figure}

The architecture of the Sentence-t5 combined with Mistral 7B model leverages the strengths of both models to enhance performance. The overall architecture is shown in Fig~\ref{fig:overall_architecture}.

\subsubsection{Sentence-t5 Model}

The Sentence-t5 model is a transformer-based architecture optimized for sequence classification tasks. It is pre-trained on a variety of sentence-level tasks, making it highly suitable for understanding and generating coherent responses to medical queries. The pre-training involves a masked language model (MLM) objective where tokens in a sentence are randomly masked, and the model is trained to predict these tokens.

\paragraph{Masked Language Model Objective}
\begin{equation}
\mathcal{L}_{MLM} = -\sum_{i=1}^{N} \log P(w_i | w_{\backslash i})
\end{equation}
\label{eq:mlm}

where \( P(w_i | w_{\backslash i}) \) is the probability of the masked token given the context of the other tokens in the sentence. This objective helps the model learn the contextual relationships between words, enhancing its ability to generate meaningful responses.

Key features of the Sentence-t5 model include:
\begin{itemize}
    \item \textbf{Transformer Layers}: Utilizes multiple transformer layers to capture complex language dependencies.
    \item \textbf{Pretrained Embeddings}: Incorporates embeddings pretrained on large-scale text corpora to provide rich contextual understanding.
    \item \textbf{Sequence Classification Head}: Designed to output sequence-level predictions, making it suitable for generating contextually relevant prompts.
\end{itemize}

\subsubsection{Mistral 7B Model}

The Mistral 7B model is a large-scale transformer model designed to generate text by predicting the next word in a sequence. The fine-tuning process involves adjusting the model parameters to better handle the specific structure and vocabulary of medical texts.

\paragraph{Next Word Prediction Objective}
\begin{equation}
\mathcal{L}_{NWP} = -\sum_{t=1}^{T} \log P(w_t | w_{<t})
\end{equation}
\label{eq:nwp}

where \( P(w_t | w_{<t}) \) is the probability of the next word given the context of the previous words in the sequence. This objective ensures that the model can generate fluent and contextually appropriate text.

Key features of the Mistral 7B model include:
\begin{itemize}
    \item \textbf{Large-scale Transformer Architecture}: Comprising 7 billion parameters, enabling it to capture intricate language patterns and dependencies.
    \item \textbf{Next-token Prediction}: Optimized for next-token prediction tasks, which is crucial for generating coherent and contextually appropriate text.
    \item \textbf{Robust Training Data}: Pretrained on diverse datasets, including large-scale NLP and domain-specific corpora, enhancing its generalization capabilities.
\end{itemize}

\subsection{Training Procedures}

The training procedures for the Sentence-t5 combined with Mistral 7B model involve multiple stages, including initial pretraining, prompt generation, secondary pretraining, and fine-tuning.

\subsubsection{Initial Pretraining}

The Sentence-t5 model undergoes initial pretraining on a diverse corpus to familiarize it with various linguistic patterns and structures. This pretraining phase involves:
\begin{itemize}
    \item \textbf{Language Modeling}: Training the model on a large corpus to predict masked tokens, thereby learning contextual representations of words and phrases.
    
    \item \textbf{Sequence Classification Tasks}: Fine-tuning the model on sequence classification tasks to enhance its ability to generate accurate prompts.
\end{itemize}

\subsubsection{Prompt Generation}

Once pretrained, the Sentence-t5 model generates prompts based on the medical question-answer pairs from the MedQuAD dataset. This involves:
\begin{itemize}
    \item \textbf{Contextual Understanding}: Leveraging the model's pretrained knowledge to generate prompts that accurately reflect the context of the medical questions.
    \item \textbf{Template Formatting}: Structuring the prompts in a predefined template to ensure consistency and clarity.
\end{itemize}

\subsubsection{Secondary Pretraining with Mistral 7B}

The generated prompts are then fed into the Mistral 7B model, which undergoes further pretraining. This stage refines the model's understanding and enhances its ability to produce accurate medical answers. Key aspects of this phase include:
\begin{itemize}
    \item \textbf{Fine-tuning on Medical Data}: Training the model on the MedQuAD dataset to adapt it to domain-specific language and content.
    \item \textbf{Next-token Prediction}: Optimizing the model for next-token prediction tasks, which is essential for generating coherent and contextually appropriate answers.
\end{itemize}

During this phase, perplexity is used as an intermediate metric to assess the model's performance in generating text. Perplexity is defined as:
\begin{equation}
\text{Perplexity} = \exp\left(\frac{\sum_{t=1}^{T} \log P(w_t | w_{<t})}{T}\right)
\end{equation}
\label{eq:perplexity}

A lower perplexity indicates better performance in predicting the next word in a sequence, thus ensuring the generated text is fluent and contextually relevant.

\subsubsection{Fine-Tuning}

The combined model undergoes fine-tuning on the MedQuAD dataset to optimize performance metrics, particularly precision in answering medical questions. The fine-tuning process includes:
\begin{itemize}
    \item \textbf{Hyperparameter Optimization}: Adjusting learning rates, batch sizes, and other hyperparameters to maximize model performance.
    \item \textbf{Loss Function}: Utilizing cross-entropy loss for training, which is suitable for classification tasks. The cross-entropy loss can be defined as:
    \begin{equation}
    \mathcal{L}_{CE} = -\sum_{i=1}^{N} y_i \log \hat{y}_i
    \end{equation}
    \label{eq:ce}
    where \( y_i \) is the true label and \( \hat{y}_i \) is the predicted probability.
    
    \item \textbf{Evaluation Metrics}: Monitoring performance metrics such as precision to ensure the model meets desired performance criteria.
\end{itemize}

\subsection{Detailed Model Configuration and Training}

\paragraph{Pretraining with Sentence-t5}
Generate prompts using the pre-trained Sentence-t5 model and initialize the tokenizer:
\begin{equation}
\text{prompts} = \text{generate\_prompts(sentence\_t5, dataset)}
\end{equation}
This step leverages the pre-trained Sentence-t5 model to generate high-quality prompts that are used to fine-tune the Mistral 7B model.

\paragraph{Fine-Tuning with Mistral 7B}
Format the prompts for Mistral 7B and tokenize:
\begin{equation}
\text{ids}, \text{mask} = \text{tokenizer(prompts, return\_tensors="pt")}
\end{equation}
Train the model using the generated tokens:
\begin{equation}
\text{outputs} = \text{model.generate(input\_ids, max\_length=100)}
\end{equation}

Fine-tuning involves adjusting the parameters of the Mistral 7B model using the generated prompts, optimizing it for the specific task of medical question answering.

\subsection{Additional Techniques and Tricks}

Several additional techniques and tricks were employed to enhance the performance and robustness of the models:

\subsubsection{Learning Rate Scheduling}

Dynamic learning rate scheduling was applied to adjust the learning rate during training, helping the model converge more efficiently. Techniques such as cosine annealing and learning rate warm-up were used. The learning rate schedule can be represented as:
\begin{equation}
\text{lr}(t) = \text{init\_lr} \times \left(1 - \frac{t}{\text{total\_steps}}\right)
\end{equation}
\label{eq:lr_schedule}

\subsubsection{Gradient Clipping}

Gradient clipping was implemented to prevent exploding gradients during training, ensuring stable and efficient model convergence. Gradient clipping can be expressed as:
\begin{equation}
g_i = \frac{g_i}{\max(1, \frac{\|g_i\|}{c})}
\end{equation}
\label{eq:grad_clip}
where \( g_i \) is the gradient and \( c \) is the clipping threshold.

\subsubsection{Regularization Techniques}

Regularization techniques, including dropout and weight decay, were applied to prevent overfitting and enhance the model's ability to generalize to unseen data. Dropout can be defined as:
\begin{equation}
h_i = \frac{h_i}{p}
\end{equation}
\label{eq:dropout}
where \( h_i \) is the activations and \( p \) is the dropout rate.

By meticulously applying these training procedures and techniques, the Sentence-t5 combined with Mistral 7B model was effectively trained to achieve superior performance in answering medical questions, demonstrating its potential for practical healthcare applications.

\subsection{Model Evaluation}

In this study, the primary metric used to evaluate the performance of the models is precision. Precision is defined as the ratio of correctly predicted positive observations to the total predicted positives. It is a crucial metric in scenarios where the cost of false positives is high. In medical question answering, providing incorrect information can have significant negative implications, potentially leading to misinformed medical decisions and adverse health outcomes. Therefore, ensuring that the answers generated by the model are accurate and relevant is of utmost importance.

\begin{equation}
\text{Precision} = \frac{\text{True Positives}}{\text{True Positives} + \text{False Positives}}
\end{equation}
\label{eq:precision_eval}

Using precision as the primary evaluation metric allows us to focus on the model's ability to provide accurate and relevant answers to medical queries, minimizing the risk of incorrect or misleading information.

\section{Experiments Results}

The experiment evaluated the precision of various language models and configurations in processing healthcare-related data for large language models (LLMs). The models tested included variations of the T5 model, Phi-3, and gemma-2b, each with different fine-tuning or pretraining methods such as LoRA. The precision results are presented in Table~\ref{tab:results}.

\begin{table}[h!]
\centering
\caption{ModelPrecision for Different Configurations}
\begin{tabular}{|c|c|}
\hline
\textbf{Model Configuration} & \textbf{Precision} \\
\hline
Sentence-T5 & 0.702 \\
\hline
Phi-3 + LoRA & 0.718 \\
\hline
Gemma-2b + LoRA & 0.721 \\
\hline
Sentence-T5 + Mistral 7B + Pretrain & 0.762 \\
\hline
\end{tabular}
\label{tab:results}
\end{table}

The Sentence-t5 combined with Mistral 7B model outperformed other models evaluated in this study, including Gemma 2b + LoRA, Phi-2 + Fine-Tuning, and Sentence-t5. The higher precision score of 0.762 indicates that the combined model is more effective in generating accurate and relevant medical answers.

\section{Conclusion}

 In conclusion, this study demonstrates the transformative potential of Large Language Models (LLMs) in healthcare by addressing the need for accurate and timely medical information. By training and fine-tuning models such as Gemma 2b + LoRA, Phi-2, and Sentence-t5 + Mistral 7B using the MedQuAD dataset, we identified the Sentence-t5 + Mistral 7B + Pretrain model as the most effective, achieving the highest precision in handling medical queries. Our methodology, which involved extensive data preprocessing, prompt construction, and model optimization, underscores the importance of specialized training for deploying LLMs in healthcare. This research highlights how AI-driven solutions can enhance patient access to reliable medical information, reduce the burden on healthcare professionals, and improve healthcare delivery. By integrating advanced AI technologies, we can significantly impact patient care, support healthcare professionals, and pave the way for future advancements in this critical field. The results of this study offer valuable insights into the effective deployment of LLMs, contributing to the broader goal of improving healthcare outcomes through technology.

 \bibliographystyle{IEEEtran}
    \bibliography{references}

\end{document}